\documentclass[9pt]{extarticle}
\usepackage{spconf}
\usepackage{amsmath}
\usepackage{amssymb}
\usepackage{caption}
\usepackage{hyperref}
\usepackage{url}
\usepackage{lipsum}
\usepackage{outlines}
\usepackage{multirow}
\usepackage{booktabs}
\usepackage{float}
\usepackage[final]{graphicx}

\usepackage{soul}
\usepackage{color}

\usepackage[normalem]{ulem}

\newcommand\blfootnote[1]{%
  \begingroup
  \renewcommand\thefootnote{}\footnote{#1}%
  \addtocounter{footnote}{-1}%
  \endgroup
}

\title{Feedback Recurrent AutoEncoder}

\name{
Yang Yang~$^{ \dagger}$, 
Guillaume Sauti\`ere~$^{ \star}$, 
J. Jon Ryu~$^{\dagger\ddagger}$, 
Taco S Cohen~$^{\star}$
}
\address{
$^{\dagger}$ Qualcomm AI Research, Qualcomm Technologies, Inc.\\
$^{\star}$ Qualcomm AI Research, Qualcomm Technologies Netherlands B.V.\\
$^{\ddagger}$ Department of Electrical and Computer Engineering, University of California San Diego}

\begin{document}
\maketitle

\begin{abstract}

In this work, we propose a new recurrent autoencoder architecture, termed Feedback Recurrent AutoEncoder (FRAE), for online compression of sequential data with temporal dependency. The recurrent structure of FRAE is designed to efficiently extract the redundancy along the time dimension and allows a
compact discrete representation of the data to be learned. 
We demonstrate its effectiveness in speech spectrogram compression.
Specifically, we show that the FRAE, paired with a powerful neural vocoder, can produce high-quality speech waveforms at a low, fixed bitrate.
We further show that by adding a learned prior for the latent space and using an entropy coder, we can achieve an even lower variable bitrate.  
\end{abstract}

\begin{keywords}
recurrent autoencoder, lossy compression, speech coding, prior model 
\blfootnote{Qualcomm AI Research is an initiative of Qualcomm Technologies, Inc. and/or its subsidiaries}
\end{keywords}

\section{Introduction}
\label{sec:intro}

Autoencoders and variational autoencoders \cite{Kingma2013} represent an important family of models commonly used for representation learning, whose main goal is to derive a compact encoding of the data that explains its underlying structure in an unsupervised way. This compact encoding lends itself to a wide variety of use cases including forming a better representation of the data, to be used as features for downstream supervised tasks, or lossy data compression after discretization. In this work, we focus on the data compression use case.

While it is natural to use a forward-only encoder and decoder architecture for non-sequential data like images \cite{Mentzer2018}, there is no standard autoencoder architecture for temporally correlated data  that has variable-length and long range dependencies such as video, speech, and text. The main challenge lies in the difficulty in capturing correlation information at different time-scales in an online/sequential fashion.

Among many existing works that apply auto-encoding to sequential data, \cite{Fabius2014, Sutskever} use an RNN for both encoder and decoder, where the encoder summarizes the entire input sequence into the last recurrent state, which is then used as the recurrent state initialization for a decoder. This approach would not work well with variable input length, as a fixed-length code is used irrespective of the sequence length. \cite{Li2018} proposes a two-scale autoencoder design, whereby a local-scale encoder extracts dynamic information of the data in each time step, and a global-scale one encodes long-term information common to the entire sequence. This two-scale design, although providing a nice disentanglement of global and local information, is not suitable for online compression setting  since the global encoding is not available until the entire sequence is observed. Another approach is to divide the long sequence into blocks and conduct auto-encoding on each block, e.g., as is done in \cite{Habibian2019}, where the video data is divided into blocks of 8 frames. The drawback is that chunking creates discontinuity in between blocks and hinders the learning of temporal dependencies with range longer than the block size.  In VQ-VAE \cite{Oord2017, Garbacea2019}, the authors apply a sequential encoder and a WaveNet type of decoder for audio data, whereas we show that decoupled encoder and decoder without feedback may lead to sub-optimal performance.

In this paper, we propose a new autoencoder architecture tailored for learning a compact discrete representation of  temporally correlated data in a sequential fashion. The discrete bottleneck codes can be used as lossy source codes of the sequential data. 

We demonstrate the power of the proposed architecture for speech spectrogram compression, where the data at each time step is a frame of a spectrogram. 
Since spectrograms (and its variants such as cepstrum) are often used as speech features in automatic speech recognition (ASR) and neural waveform synthesis \cite{Valin2019, Wang2019, Prenger2019}, an efficient compression could help reduce bandwidth in cloud-based ASR as well as bitrate in neural-vocoder based speech coding.

The contributions of this work are as follows. (i) We propose a new recurrent autoencoder architecture, termed feedback recurrent autoencoder (FRAE), with the salient feature of decoder-to-encoder recurrent state feedback, which is shown to be superior to other recurrent schemes. (ii) The recurrent structure of FRAE facilitate an easy extension to its variational counter-part that allow variable rate encoding. (iii) We show that the system can produce high-quality speech waveforms at a low bitrate when paired with a powerful neural vocoder.

\newcommand{\xb}{\mathbf{x}}
\newcommand{\zb}{\mathbf{z}}

\begin{figure*}[ht]
\begin{center}
\includegraphics [width=0.95\textwidth]{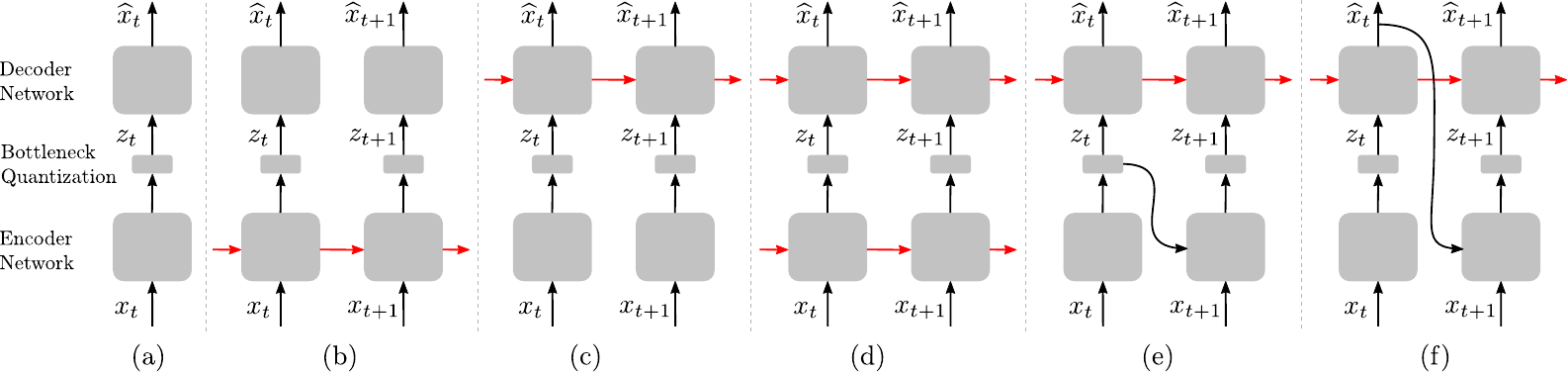}
\vspace{-0.25cm}
\captionsetup{width=.85\linewidth, justification=centering}
\caption{Different recurrent autoencoder schemes (recurrent connections are displayed in red) \\(a) No recurrency (b) Encoder only (c) Decoder only (d) Separate (e) Latent feedback (f) Output feedback}
\vspace{-0.25cm}
\label{fig:recurrencies}
\end{center}
\end{figure*}

\vspace{-0.2cm}
\section{Recurrent AutoEncoder}
\label{sec:rae}

In this section, we discuss and compare different autoencoding schemes for the compression of a correlated sequence $(x_t)_{t\in\mathbb{N}}$.

The most naive approach is to encode and decode data at each time-step independently, as illustrated in Fig.~\ref{fig:recurrencies}(a). Since the network operations at different time-steps are decoupled, however, any temporal correlation is completely ignored and thus the latent codes $(z_t)_{t\in\mathbb{N}}$ necessarily encode redundant information over time, leading to inefficient compression/representation of the correlated data sequence.

Hence, to capture any temporal correlation, we require time-domain coupling in the autoencoding process, which can be achieved by either (1) applying a feed-forward network that can access data from multiple time-steps, e.g. using temporal convolution \cite{Garbacea2019} or self-attention \cite{Vaswani2017},
or (2) using recurrent network architectures. In this work, we focus on the latter approach. We list several designs of recurrent structures in Fig. 1(b)-(f) and identify defects in each design to motivate the proposed architecture.

First, adding recurrent connection to only the encoder or the decoder, as shown in Fig.~\ref{fig:recurrencies}(b) and (c), does not allow the latent codes to utilize temporal correlation and thus is inherently flawed.  
In case (b), the encoder can access history information of input data through the recurrent connection, while the decoder only has access to $z_t$ for each time-step $t$. As a result, each $z_t$ is supposed to fully describe $x_t$, and thus temporal redundancy is not being utilized.
Similarly for case (c), adding the recurrent connection on the decoder alone has little benefit given that each latent code $z_t$ is formed only from the current data input $x_t$.

It is also a bad design to simply add separate recurrent connections to  the encoder and decoder, as illustrated in Fig.~\ref{fig:recurrencies}(d). In this case, even though both encoder and decoder have access to history information, due to limited dimension in the bottleneck, the information that the decoder has access to is a lossy version of that exposed to the encoder. This is even more pronounced when bottleneck latents are quantized into a finite set of discrete values to perform lossless compression of latent codes. Due to this mismatch, the encoder is unable to construct latent codes that are tailored for decoder's context. This motivates a feedback connection from decoder to encoder.

For the two schemes illustrated in Fig.~\ref{fig:recurrencies}(e) and (f), either the code $z_{t-1}$ or the reconstructed output $x_{t-1}$ is fed back to the encoder at time-step $t$. These feedback connections inform the encoder of the decoder status at the previous time-step, ignoring longer range dependencies. For case (f), we empirically observe instability in training. 

It is worth noting that the video compression framework of DVC \cite{Lu2018} can be viewed as an instantiation of Fig.~\ref{fig:recurrencies}(f) where decoded data in the previous time steps are fed back to the encoder for explicit motion and residual information compression, and the one proposed in VQ-VAE \cite{Oord2017, Garbacea2019} can be viewed as a convolutional variant of Fig.~\ref{fig:recurrencies}(d) where both encoder and decoder use convolution to cover a large temporal receptive field without any decoder-to-encoder feedback.

\subsection{Feedback Recurrent AutoEncoder}
\begin{figure}[!b]
\begin{center}
\includegraphics [width=0.45\textwidth]{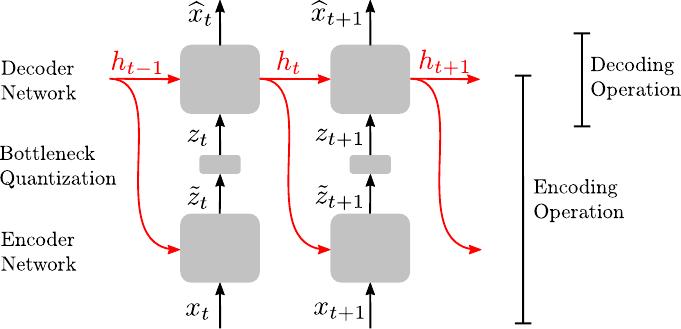}
\caption{Feedback Recurrent Autoencoder (FRAE)
}
\label{fig:frae}
\end{center}
\end{figure}

Based on the above discussion, an ideal sequential autoencoding design should have the following properties:
(1) both encoder and decoder have recurrent connections; 
(2) there is feedback from decoder to encoder;
(3) capable of utilizing long-term temporal correlation.

We introduce a simple autoencoder design that satisfies all three requirements. As illustrated in Fig.~\ref{fig:frae}, the salient feature is that the decoder feeds back its recurrent state to the encoder, hence the name \emph{Feedback Recurrent AutoEncoder} (FRAE). 
This structure can be interpreted as a \emph{non-linear predictive coding} scheme: the recurrent state $h_t$ contains a summary of previously decoded frames; the encoder may take advantage of the existing information in $h_t$ to form a code $z_{t+1}$ which encodes only the residual information missing to reconstruct $x_{t+1}$ from $h_t$. We can interpret $h_{t}$ as containing information regarding the prediction/extrapolation of the next frame $x_{t+1}$, corrected by additional/residual information from the encoder captured in $z_{t+1}$.

The proposed FRAE architecture can be used in online lossy compression of sequential data at a fixed-rate: at each time step $t$, encoder converts a data sample $x_t$ to the quantized latent code $z_t$, and then losslessly transmits $z_t$ to decoder with a fixed-length code.
The bitrate of this scheme is determined by the dimension of the bottleneck and the size of quantization alphabet in each dimension.
Note that the encoding process requires running the decoder network up to the generation of the decoder recurrent state, as illustrated in Fig.~\ref{fig:frae}. This resembles the \emph{analysis-by-synthesis} principle, as the encoding process carries out decoding operations to come up with a better code.

In Table~\ref{table:recurrency_schemes}, we compare the performance of different recurrent schemes when applied to speech spectrogram compression. The results show that FRAE leads to lower reconstruction distortion compared with any other scheme in Fig.~\ref{fig:recurrencies} with the same bottleneck dimension and quantization scheme. %Details of the experiments will be presented in Section~\ref{subsec:recurrencies}. 
Details of the experiment settings are deferred to Section~\ref{subsec:recurrencies}.

\begin{table}[ht]
\caption{Test set performance of different recurrent schemes for wide-band (16KHz) spectrogram compression at 1.6Kbps.}\label{table:recurrency_schemes}\vspace{0.2cm}
\centering
\begin{tabular}{ @{}l c c c@{} } 
\toprule
&& \multicolumn{2}{c}{POLQA score}\\
\cmidrule{3-4}
Recurrency & Mel-scale & original  & Griffin-Lim  \\
scheme     & MSE       & phase     & 100-iter \cite{Griffin1984} \\
\midrule
(a) No recurrency & 18.369 & 2.404 & 1.489\\ 
(b) Encoder only & 18.343 & 2.507 & 1.566\\ 
(c) Decoder only & 16.389 & 3.091 & 1.878\\ 
(d) Separate& 14.283 & 3.588 & 2.174\\ 
(e) Latent feedback & 14.475 & 3.589 & 2.131\\ 
(f) Output feedback & 13.915 & 3.594 & 2.159\\ 
FRAE & {\bf 13.003} & {\bf 3.929} & {\bf 2.350}\\ 
\bottomrule
\end{tabular}
\end{table}

Like FRAE, the DRAW architecture \cite{Gregor2015, Gregor2016} also features a feedback connection from the decoder to the encoder. Its motivation and application, however, is different from FRAE: DRAW focuses on the progressive encoding of a single image, using feedback to allow the network to correct its previous mistakes in an iterative fashion, while FRAE focuses on online compression of sequential data and uses feedback as a way to extract long-term temporal redundancy and provide complementary information to the decoder.

\subsection{Feedback Recurrent Variational AutoEncoder}
\label{subsec:frvae}
The average bitrate of the previously described lossy compression scheme may be further reduced without sacrificing a distortion level, using a variable-length code with a trainable probability model $p_{\text{prior}}(\zb)$ over the latent code $\zb=(z_1,\ldots,z_T)$.
We refer to this combination of FRAE and the latent probability model as feedback recurrent \emph{variational} autoencoder (FR-VAE).

To train a FR-VAE model, we train the FRAE architecture and the probability model $p_{\text{prior}}(\zb)$ jointly based on the following objective that quantifies the rate-distortion of the scheme:
\begin{align}\label{eq:rate_distortion_objective}
\sum_{t=1}^T d(x_t,\hat{x}_t) + \beta \log\frac{1}{p_{\text{prior}}(z_t|z_{<t})},
\end{align}
where $z_t\triangleq \text{encoder}(x_t,h_{t-1})$, $\hat{x}_t\triangleq \text{decoder}(z_t,h_{t-1})$, and $d(x,\hat{x})$ denotes a distortion function.
The first term captures the amount of distortion incurred, while the second term captures the \emph{rate} as the amount of the ideal codeword length of $z_t$ when $p_{\text{prior}}(z_t|z_{<t})$ is used for entropy coding. 
We can trade-off between the rate and the distortion by sweeping the hyper-parameter $\beta> 0$. The regular FRAE training is a special case with $\beta=0$, or it can be viewed as having a fixed uniform prior on the latent codes. We remark that this this rate--distortion objective in Eq.~\eqref{eq:rate_distortion_objective} is equivalent to $\beta$-VAE objective if the encoder is deterministic as in our case; we refer the interested readers to \cite{Ghosh2019, Habibian2019} for a detailed discussion on lossy compression and variational inference.

\begin{figure}[ht]
\begin{center}
\includegraphics [width=0.33\textwidth]{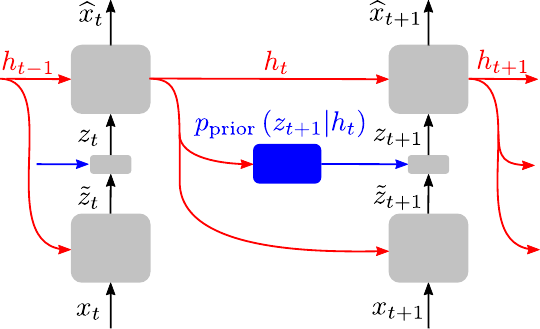}
\caption{Feedback Recurrent Variational AutoEncoder (FR-VAE) }
\vspace{-0.3cm}
\label{fig:frvae}
\end{center}
\end{figure}

We propose to use a prior model $p_\text{prior}(z_t|h_{t-1})$ for an autoregressive prior $p_{\text{prior}}(z_t|z_{<t})$, as the decoder recurrent state $h_{t-1}$ already summarizes the history of the latent code $\zb_{< t}$; see Fig.~\ref{fig:frvae}.
Not only it requires only a small add-on to the existing FRAE architecture, but we also empirically demonstrate that this specific design choice leads to a better rate-distortion trade-off compared with a time-invariant prior model $p_\text{prior}(z_{t})$ and a prior model $p_\text{prior}(z_{t}|z_{t-1})$ that is conditioned only on the latent codes from the previous time-step; see Section~\ref{subsec:priors} for detailed experiment.

\section{Experiments}
\label{sec:experiments}

In this section, we focus on the problem of speech spectrogram compression and conduct three experiments. 
In Section~\ref{subsec:recurrencies} we demonstrate the effectiveness of FRAE by comparing its performance with other recurrent autoencoder designs in Fig.~\ref{fig:recurrencies}. 
In Section~\ref{subsec:priors}, we focus on FR-VAE and compare the rate-distortion trade-off of different prior models. In Section~\ref{subsec:wavenet}, we use FRAE transcoded speech spectrograms to condition a WaveNet \cite{Oord2016} and generate high-quality speech waveforms at a low bitrate. 

Across all experiments, recurrent autoencoder networks are constructed using a combination of convolutional layers, fully-connected layers, and GRU layers, with a total of around 1.5 million parameters. For bottleneck quantization, we apply the technique used in \cite{Habibian2019} to quantize each dimension independently with a jointly learned codebook of size four. 

Regarding datasets, for the first two experiments, we use  LibriVox audiobook recording of Agnes Grey, a studio-quality single speaker dataset, with 2.3 hours for training and 13 minutes for testing. For the last experiments with WaveNet, we use the multi-speaker WSJ1 dataset, which has 66 hours from 200 speakers for training and 2.2 hours with 10 speaker for testing. The train/test split has a disjoint set of speakers and utterances with even gender distribution. Wide-band (16KHz sampling rate) audios are used for both dataset. Each data sample $\mathbf{x}$ is the spectrogram of a speech clip, with $x_t$ representing a single frame of spectrogram at dB scale. The spectrograms are computed from square-root Hanning windowed STFT with window shift of 160 (10ms) and window size (same as FFT-size) of 320 (20ms), corresponding to a frame rate of 100Hz.

A Mel-scale mean squared error (Mel-scale MSE) is used as the reconstruction loss $d(x_t, \hat{x}_t)$ for training, where the MSE of each frequency bin is scaled according to its weight at Mel-frequency \cite{Slaney1998} to capture human perceptual sensitivity with respect to frequencies. Specifically, the weight on frequency $f$ is defined as
\begin{align}
w[f] = \left\{
\begin{array}{ll}
1 & f\leq 1000\text{ Hz}\\
969.672/f & f > 1000\text{ Hz}
\end{array}
\right.\notag.
\end{align}

\subsection{Comparison of different recurrency schemes}
\label{subsec:recurrencies}

In this experiment, we compare the performance of the recurrency autoencoding schemes listed in Fig.~\ref{fig:recurrencies} with FRAE. We fix the bottleneck dimension to be 8 for all these autoencoder schemes. Given that each bottleneck dimension has 4 quantization levels and the frame-rate of spectrogram is 100Hz, the spectrogram is compressed at a fixed bitrate of 1.6Kbps. 

Aside from the Mel-scale MSE, we also evaluated the models by converting the transcoded spectrogram back to the time-domain by Inverse-STFT, where we then compute the POLQA score \cite{polqa}, an objective perceptual metric of audio quality, with respect to the ground-truth waveform. The phase of the spectrogram is either taken from the original (genie phase) or computed by running Griffin-Lim algorithm \cite{Griffin1984} for 100 iterations. The performance comparison is detailed in  Table~\ref{table:recurrency_schemes}. For all three metrics, FRAE outperforms the rest with a significant margin, which demonstrates the effectiveness of the feedback recurrent design.

It is worth mentioning that even though the output feedback scheme in Fig.~\ref{fig:recurrencies}(f) achieves the second best performance, in practice we find it is prone to divergence during training and thus hard to optimize, which may be attributed to the fact that it increased the depth of computation graph after RNN unrolling without proper gating mechanism (as that in GRU or LSTM) to alleviate gradient explosion problem. In constrast, the FRAE scheme always leads to stable training.

\vspace{-0.1cm}
\subsection{Comparison of different prior models of FR-VAE}
\setcounter{footnote}{-1} 
\label{subsec:priors}

\begin{figure}[h]
\begin{center}
\includegraphics[width = 0.39\textwidth]{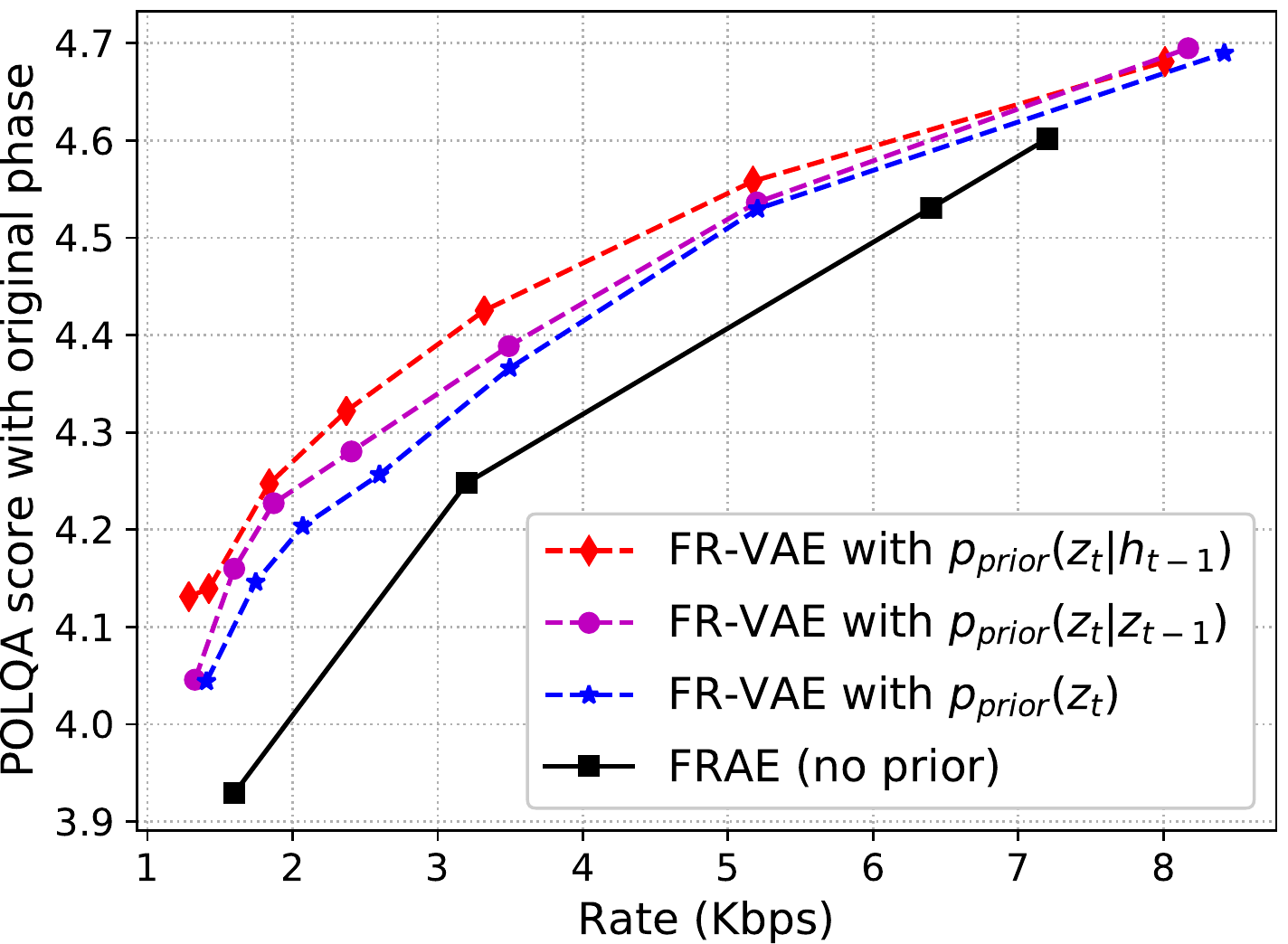}
\medskip\\
\caption[]{Rate-distortion of different prior models \footnotemark. Distortion is represented by the average POLQA score of test set after converting autoencoded spectrogram to waveform using the original phase.}
\label{fig:rd_prior}
\vspace{-0.3cm}
\end{center}
\end{figure}

Next, we train FR-VAE with three varients of prior models: one that is conditioned on $h_{t-1}$ as illustrated in Fig.~\ref{fig:frvae}, one that is conditioned on $z_{t-1}$; and a time-invariant model without any conditioning. A simple MLP is used for the first two. The rate-distortion trade-offs are shown in Fig.~\ref{fig:rd_prior}, with the distortion of each model represented by the average POLQA score of the waveforms generated using the autoencoded spectrogram together with the original phase.
\footnotetext{The birates of FR-VAE in Fig.~\ref{fig:rd_prior} were computed as the second term in Eq.~\eqref{eq:rate_distortion_objective} without the $\beta$ scaling. If we assume the use of adaptive arithmetic coding as the entropy coding scheme, then in practice there can be at most 2-bit overhead per codeword, which translates into 50bps overhead in bitrate assuming that each codeword encodes 4 consecutive frames (40ms) of spectrogram.}

The three prior models are trained with bottleneck size of 48 and sweeping of $\beta$ from $0.001$ to $0.007$ with step size of $0.001$, optimized for Eq.~\eqref{eq:rate_distortion_objective}. The results are compared with FRAE (fixed, uniform prior) with bottleneck size of 8, 16, 32, and 36, corresponding to a fixed bitrate of 1.6, 3.2, 6.4, 7.2~Kbps, respectively. From the results we can see that $p_\text{prior}(z_t|h_{t-1})$ as proposed in Fig.~\ref{fig:frvae} achieves the best rate-distortion among the four with a maximum rate reduction of around 1.5~Kbps compared to FRAE.

\subsection{Waveform generation using WaveNet as phase model}
\label{subsec:wavenet}

In this experiment, we pair FRAE with a WaveNet model to generate speech waveform. Four FRAE models are trained with the same configurations as the previous experiment (1.6, 3.2, 6.4, 7.2~Kbps). We then freeze the FRAE models and train four separate WaveNets, each conditioned on the autoencoded spectrogram from one of the FRAE models. Speaker identity is not used.

\begin{figure}[!h]
\begin{center}
\includegraphics [width=0.37\textwidth]{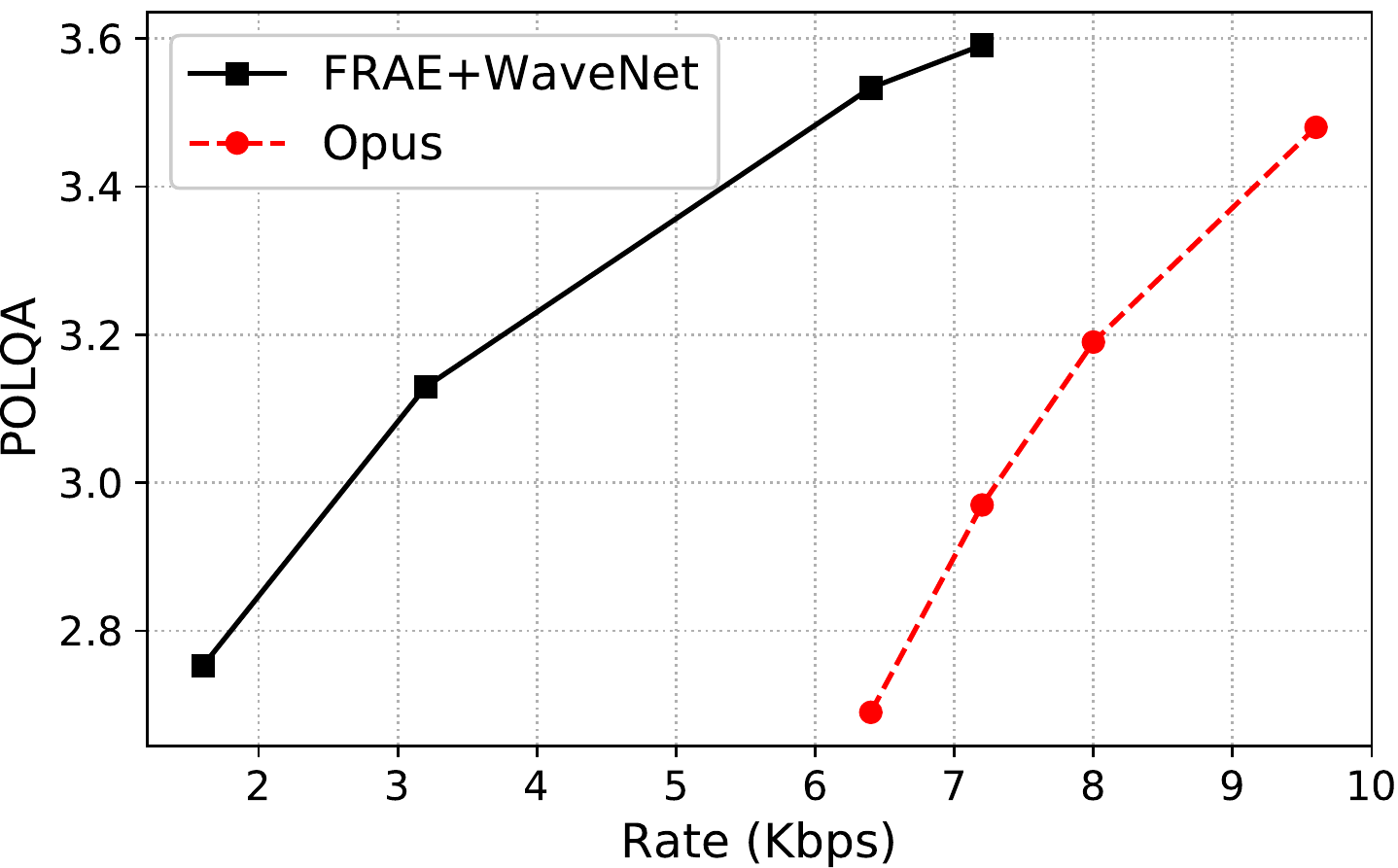}
\caption{
POLQA score vs bitrate for FRAE+WaveNet, trained on WSJ1 and evaluated on WSJ1 test set, against Opus.
}
\vspace{-0.5cm}
\label{fig:wavenet_wsj}
\end{center}
\end{figure}

\begin{figure}[!h]
\begin{center}
\includegraphics [width=0.37\textwidth]{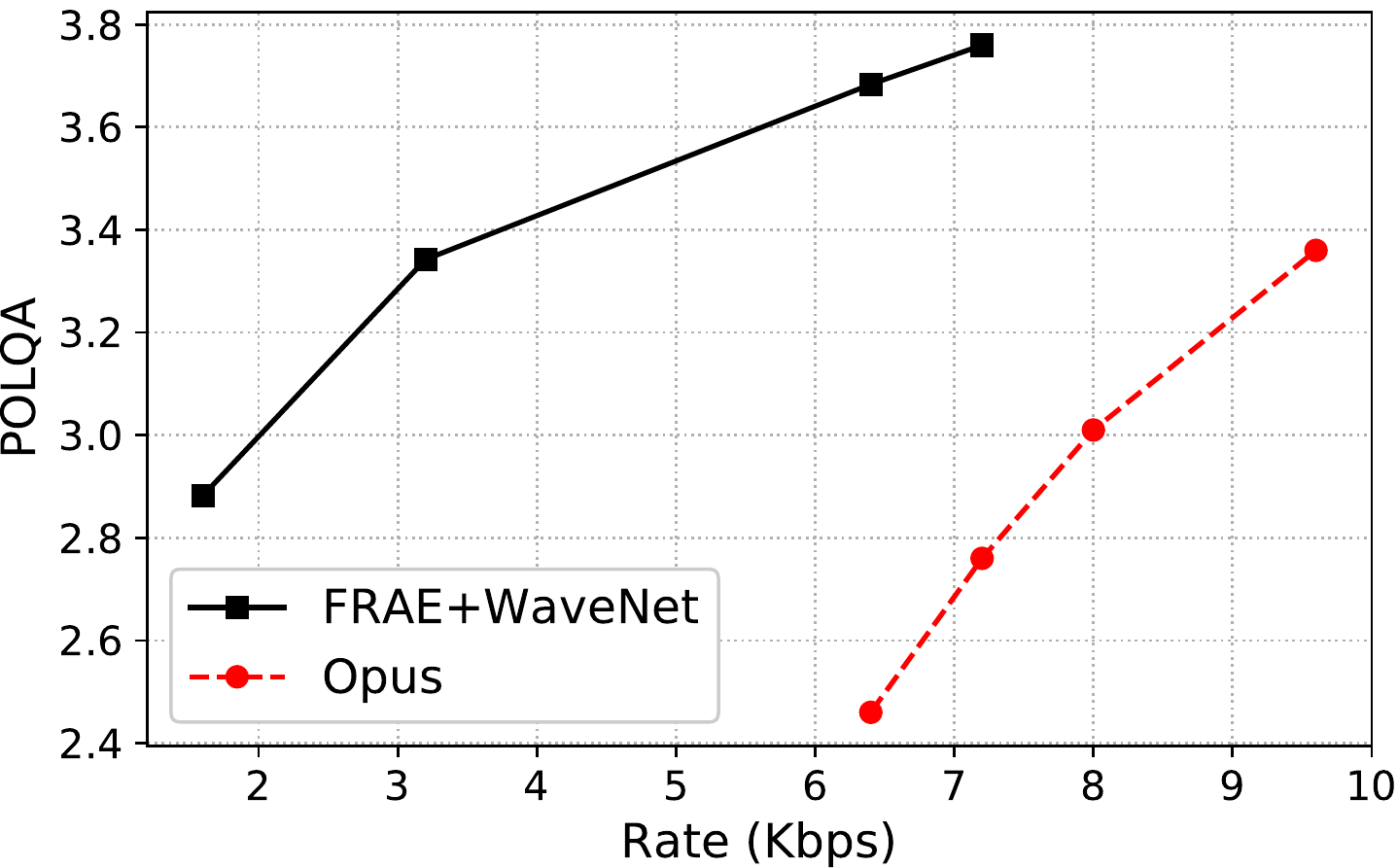}
\caption{
POLQA score vs bitrate for FRAE+WaveNet, trained on LibriVox-Agnes-Grey and evaluated on LibriVox-Agnes-Grey test set, against Opus.
}\label{fig:wavenet_audiobook}
\vspace{-0.3cm}
\end{center}
\end{figure}

The experiment is conducted separately on the the multi-speaker WSJ1 dataset and the LibriVox-Agnes-Grey audiobook dataset, with the average POLQA score shown in Fig.~\ref{fig:wavenet_wsj} and Fig.~\ref{fig:wavenet_audiobook} respectively and compared against Opus \cite{opus}, a well-known open source speech codec. It can be seen that significant bitrate reduction is achieved by FRAE+WaveNet when compared with Opus, and the gap widens as the bitrate goes lower.

\section{Conclusion}
\label{sec:conclusion}

In this work we presented a new scheme of recurrent autoencoder in the context of lossy online compression of temporally correlated data, and demonstrated its effectiveness on the speech spectrogram compression task. We showed that high-quality waveform can be generated at a low bitrate when it is used together with WaveNet, and the bitrate can be reduced further by adding a prior model on the latent codes. An interesting future direction is to modify the FRAE architecture by allowing multiple update rates for different part of the latent codes,capturing correlation information at different time-scales, to achieve further reduction in bitrate.

\clearpage
\bibliographystyle{IEEEbib}
\bibliography{refs}

\end{document}